\documentclass{article}

\usepackage{PRIMEarxiv}

\usepackage{graphicx} 
\usepackage{subfig}
\usepackage{epsfig} 
\usepackage{mathptmx} 
\usepackage{times} 
\usepackage{amsmath} 
\usepackage{amssymb}  
\usepackage{tipa}
\usepackage{cite}
\usepackage{array}
\usepackage{hyperref}
\usepackage{fixltx2e}
\usepackage{color}
\usepackage{dblfloatfix}
\usepackage{url}

\usepackage{tikz}
\usepackage{algorithm}
\usepackage{algorithmic}
\usepackage{booktabs}
\usepackage{float}
\usepackage{amsmath,amssymb,amsfonts}
\usepackage{mathrsfs}
\usepackage{soul,color}
\usepackage{graphicx}
\graphicspath{{figs/}}

\usepackage{lipsum}
\usepackage[utf8]{inputenc}
\usepackage{multicol,tabularx,capt-of}
\usepackage{multirow}
\usepackage{nicematrix}

\title{Incorporating Target Vehicle Trajectories Predicted by Deep Learning Into Model Predictive Controlled Vehicles}

\author{
  Ni~Dang \\
  Technical University of Munich \\
  Munich, Germany\\
  \texttt{ni.dang@tum.de} \\
  \And
  Zengjie~Zhang \\
  Eindhoven University of Technology \\
  Eindhoven, Netherlands\\
  \texttt{z.zhang3@tue.nl} \\
   \And
  Jizheng~Liu \\
  Beijing Institute of Technology \\
  Beijing, China \\
  \texttt{liujz\_ev@bit.edu.cn} \\
  \And
  Marion~Leibold \\
  Technical University of Munich \\
  Munich, Germany \\
  \texttt{marion.leibold@tum.de} \\
  \And
  Martin~Buss \\
  Technical University of Munich \\
  Munich, Germany \\
  \texttt{mb@tum.de} \\
}

\begin{document}

\maketitle
\thispagestyle{empty}
\pagestyle{empty}
\begin{abstract}
Model Predictive Control (MPC) has been widely applied to the motion planning of autonomous vehicles. An MPC-controlled vehicle is required to predict its own trajectories in a finite prediction horizon according to its model. Beyond this, the vehicle should also incorporate the prediction of the trajectory of its nearby vehicles, or target vehicles (TVs) into its decision-making. The conventional trajectory prediction methods, such as the constant-speed-based ones, are too trivial to accurately capture the potential collision risks. In this report, we propose a novel MPC-based motion planning method for an autonomous vehicle with a set of risk-aware constraints. These constraints incorporate the predicted trajectory of a TV learned using a deep-learning-based method. A recurrent neural network (RNN) is used to predict the TV's future trajectory based on its historical data. Then, the predicted TV trajectory is incorporated into the optimization of the MPC of the ego vehicle to generate collision-free motion. Simulation studies are conducted to showcase the prediction accuracy of the RNN model and the collision-free trajectories generated by the MPC.


\end{abstract}

\section{INTRODUCTION}
Model Predictive Control (MPC) has attracted increasing attention in autonomous driving due to its capability of incorporating traffic rules, the physical limitations of vehicles, and the collision avoidance requirements into driving control. MPC iteratively solves an optimization problem and gets a feasible trajectory that is subject to these constraints. 
An MPC-controlled ego vehicle (EV) is said to be able to interact with a target vehicle (TV) if it can predict the future behaviors of the TV and incorporate the predicted behaviors into its decision-making, such that the risk of potential collisions is avoided. Therefore, predicting the future behaviors of a TV is an important topic to realize risk-aware autonomous driving. Trajectory prediction of an autonomous vehicle is conventionally conducted by assuming a constant speed, i.e., the TV is moving while maintaining its current speed~\cite{Tim2023journal, Dang2022, Dang2023}. However, these assumptions ignore the influence of the real-time control inputs of the TV on its future trajectory, especially when it is required to perform a different driving task in a short future horizon. To solve this problem, a more realistic prediction method that does not only consider the current state of the TV but also its historical data should be proposed to achieve precise prediction.

Other than the constant-velocity-based trajectory prediction method, learning-based methods have been used to predict the trajectory of the target vehicles based on their historical trajectories. In~\cite{Mozaffari2022}, deep learning (DL) methods have been successfully used for predicting the behaviors of vehicles. Since the vehicle trajectories can be recognized as the sequences of vehicle positions, recurrent neural network (RNN) is most used due to their capability of handling data sequences. In~\cite{hou2019interactive}, an RNN model with long-short-term memory units is used for trajectory prediction. In~\cite{dong2019interactive}, the technology of meta-induction learning is used to incorporate the interaction in a multi-vehicle system. A survey of deep learning methods to solve the vehicle trajectory prediction problem can be referred to in~\cite{mozaffari2020deep}. 
In this report, we use deep learning to predict the TV's trajectory and encode it into the safety constraint of an MPC. As a result, the MPC incorporates the interaction between the EV and the TV and thus produces risk-aware collision-free motion. To facilitate the interface between deep learning and MPC, calibration of the training data is performed. We also adjust the offset of the predicted TV trajectory to avoid the prediction errors caused by this offset. 
The rest of the paper is organized as follows. Sec. \ref{mpc} introduces the MPC formulation for autonomous vehicles. Sec. \ref{TrajectoryPrediction} presents the deep learning-based prediction method. Simulation studies that validate the efficacy of the proposed method are shown in Sec. \ref{simulations}. We conclude our work and discuss the future work in Sec. \ref{conclusion}

\section{MPC Incorporating the Predicted TV Trajectory}\label{mpc}
This section presents the MPC-based motion planning framework incorporating the predicted TV trajectory. A two-vehicle system that contains an EV and a TV is considered. The EV is described using the following linearized and discretized kinematic bicycle model~\cite{Tim2023journal}, 
\begin{equation}\label{VeicleModel}
{\boldsymbol{\xi}}_{t+1}=\boldsymbol{\xi}_{0}+T\boldsymbol{f}^\text{c}\left(\boldsymbol{\xi}_{0},\boldsymbol{0}\right)+A\left(\boldsymbol{\xi}_{t}-\boldsymbol{\xi}_{0}\right)+B\boldsymbol{u}_{t}, \ t\in\mathbb{N},
\end{equation} 
where $\boldsymbol{f}^\text{c}$ is a nonlinear continuous kinematical bicycle model introduced in~\cite{Carvalho2014, Gao2010, Levinson2011, Carvalho2013},  $\boldsymbol{\xi}_t=\left(x_t,y_t,\psi_t,v_t\right)^\intercal$ is the state vector that
consists of the longitudinal position $x_t$ and lateral position $y_t$, the velocity $v_t$ and~inertial heading $\psi_t$ of the EV at time $t$, $\boldsymbol{\xi}_0$ is the initial state of the system, $\boldsymbol{u}_t=\left(a_t,\delta_t\right)^\intercal$ is the control input that comprises the acceleration $a_t$ and steering angle $\delta_t$ at time $t$, $T$ is the sampling time, and $A$, $B$ system matrices are linearized system gains calculated according to~\cite{Tim2020arXiv}. Then, an MPC controller iteratively solves the following optimal control problem at any current time $t$, 
\begin{subequations}\label{eq:mpc}
	\begin{align}
		\mathop{\min}\limits_{\pmb u} \ \  &\sum_{k=0}^{N-1}({\Vert \boldsymbol{\xi}_k - \boldsymbol{\xi}_k^{\text {ref}} \Vert}_Q^2+{\Vert \boldsymbol{u}_k \Vert}_R^2+{\Vert \boldsymbol{\xi}_N - \boldsymbol{\xi}_N^{\text {ref}} \Vert}_S^2)\label{costfunction1}\\
		\text{s. t. }\ &\boldsymbol{\xi}_{k+1}\ =\boldsymbol f^\text{d}(\boldsymbol{\xi}_0,\boldsymbol{\xi}_k, \boldsymbol{u}_k), \ k=0,1,\cdots,N,\label{systemdynamics1}\\
		&\boldsymbol{\xi}_{k\ } \in \Xi,\ k=0,1,\cdots,N,\label{statecon1}\\
		&{\boldsymbol{u}}_{k}\ \in \mathcal U,\ k=0,1,\cdots,N-1,\label{inputcon1}\\
		&\boldsymbol{\xi}_k\in\Xi_k^{\text {safe}}, k=1,2,\cdots,N,\label{safecon}
	\end{align}	
\end{subequations}
where $k$ counts from current time $t$ on, $N$ is the prediction horizon, $\boldsymbol{\xi}^{\text {ref}}_k$ is the reference trajectory to be tracked by the EV, ${\pmb u}=(\boldsymbol{u}_0,\boldsymbol{u}_{1},\cdots,\boldsymbol{u}_{N-1})^\intercal$ is the control input sequence to be solved, $Q$~$\in$~$\mathbb R^{4\times4}$, $R$~$\in$~$\mathbb R^{2\times2}$ and $S$~$\in$~$\mathbb R^{4\times4}$ are the weighting matrices, $\boldsymbol f^\text{d}$ is the EV model defined in \eqref{VeicleModel}, $\Xi$ is the safety set to describe the road boundaries, the limitations of the EV, and the traffic rules, $\mathcal{U}$ is the feasible control set, and $\boldsymbol{\xi}_k\in\Xi_k^{\text {safe}}$ is the safety set for the collision avoidance with the TV.

For any current time $t$, the safety constraint $\boldsymbol{\xi}_k\in\Xi_k^{\text {safe}}$ ensures that the EV avoids potential collisions with the TV at prediction time step $k$. In this paper, we define $\boldsymbol{\xi}_k\in\Xi_k^{\text {safe}}$ as an elliptical region around the TV. The center of the ellipse is the geometric center of the TV. The size of the ellipse is sufficiently large to cover the size of the TV. Let $x_k^{\text TV}$ and $y_k^{\text TV}$ denote the longitudinal and lateral positions of the predicted TV trajectory at step $k$. Then, the distance between the vehicles in the two directions at prediction step $k$ are $\Delta x_k = x_k-x_k^{\text TV}$ and $\Delta y_k = y_k-y_k^{\text TV}$. Then, the safety set $\Xi_k^{\text {safe}}$ is represented as
and the safety constraint is 
\begin{equation}
	\Xi_k^{\text {safe}} = \left\{ (\Delta {x_k}, \Delta {y_k}) \left| \frac{{\Delta x_k}^2}{a^2}+\frac{{\Delta y_k}^2}{b^2}\ge1. \right. \right\}. 
 \label{safecon_exp}\\
\end{equation}

We will introduce how to predict the TV trajectory $(x^{\mathrm{TV}}_k, y^{\mathrm{TV}}_k)$, $k=0,1,\cdots,N$, in the next section.

\section{Trajectory Prediction}\label{TrajectoryPrediction}

In this section, we present how to predict the lane-changing trajectories of a vehicle using its historical trajectories. We first generate the lane-changing data of a vehicle using a polynomial interpolation method. Then, we use the generated data to train a deep neural network that is applied to predicting the future trajectory of a vehicle based on its historical trajectory.

\subsection{Data Generation}

Predicting future trajectories using historical trajectories renders a regression problem. The primary step is to create the data set used to train a certain prediction model. The data set contains a cluster of historical trajectories of a vehicle as data samples. The ground truth label of each data sample is its corresponding future trajectory. The trajectories are retrieved from a typical type of lane-changing path.

\subsubsection{Lane-Changing Path Generation}

Lane-changing requires that the vehicle smoothly switches from an original lane to a target lane while maintaining a constant longitudinal velocity $v$. Therefore, we use piece-wise polynomial splines to represent the lane-changing path of a vehicle. A typical lane-changing path consists of the following three stages.
\begin{itemize}
\item \textbf{Preparation stage (I):} the vehicle prepares the lane-changing, moving along the original lane and maintaining a constant speed $v$. This stage lasts for $2\,$s.
\item \textbf{Changing stage (II):} the vehicle changes the lane, starting from the original lane at the original speed $v$ and ending at the target lane at a target speed $v$. This stage lasts for $4\,$s.
\item \textbf{Finishing stage (III):} the vehicle completes the lane-changing, moving along the target lane at speed $v$.  This stage lasts for $2\,$s.
\end{itemize}


In this sense, the trajectories of the vehicle at stages I and III are straight lines. In stage II, the trajectory of the vehicle, represented as a sequence of planar coordinates $(x,y)$, is interpolated using a third-order polynomial function
\begin{equation}
y(x)\!=\!y_0\!+\!3(y_T\!-\!y_0)\left(\frac{x\!-\!x_0}{x_T\!-\!x_0} \right)^2\!-\!2(y_T\!-\!y_0)\left( \frac{x\!-\!x_0}{x_T\!-\!x_0} \right)^3
\end{equation}
where $(x_0, y_0)$ and $(x_T, y_T)$ are the coordinates of the starting point and the ending point of the vehicle in stage II. The longitudinal coordinate $x$ is sampled at a constant sampling time $\Delta t = 0.1\,$s. The generated trajectory is sufficiently smooth with terminal conditions $\dot{x}_0 = \dot{x}_T = v$ and $\dot{y}_0 = \dot{y}_T=0$.

To ensure the diversity of the data set, we create various lane-changing paths using different velocities $v$. For both velocities, we take their values from $10\,$m/s to $40\,$m/s with a constant increment of $0.1\,$m/s. Therefore, we ultimately obtain $301$ paths with different velocity profiles. Note that the size of the paths, namely the number of sampled coordinates in a path, are different due to different velocities $v$ but the same sampling time $\Delta t$.

\subsubsection{Segmentation}

From the generated lane-changing paths of the vehicle, we create the training and test data. Each sample of the data set is the historical trajectory of the vehicle before a certain time instant, and its ground truth label is the future trajectory starting from this instant. We define that all historical and future trajectories have the same size $M=30$. In this sense, for every generated lane-changing path, we create a data item by taking a segment that contains $2M$ successive sampled coordinates. The first half of the segment forms a sample of the data, and the other half serves as the ground truth label. The first coordinate of the future trajectory noted as $(x_s, y_s)$, is referred to as the splitting point. For a path sized $N$, $N > 2M$, we obtain $N-2M$ segments, i.e., $N-2M$ labeled data samples.

\subsubsection{Calibration}

Different data samples have different splitting points which bring different offsets to the longitudinal coordinate of the samples. To eliminate the influence of these offsets on the model training, we subtract $x_s$ from all the longitudinal coordinates of the samples, which is referred to as calibration. After calibration, the splitting points of all data samples have zero longitudinal coordinates $x_s=0$.

\subsubsection{Data Splitting}

Having performed segmentation and calibration, we obtain $L=6622$ data samples. Each sample contains a historical trajectory, with the corresponding future trajectory being its ground truth label. We split the data into a training set and a test set with a ratio of $6:4$. We also randomly shuffle the data to avoid the influence of the continuity of the vehicle motion.

\subsection{Prediction Model}

In this paper, we use a recurrent neural network (RNN) model to predict future trajectories. The RNN is composed of a sequence input layer, an encoder layer, a latent feature layer, a decoder layer, and an output layer, and is implemented using the MATLAB~\textregistered Deep Learning Toolbox. The details of the network structure are introduced as follows.

\subsubsection{Sequence Input Layer}

This is a typical input layer used to feed the historical trajectories into the RNN. In MATLAB~\textregistered, it is created using function \textit{sequenceInputLayer}. The input size is 2, namely the number of planar dimensions. This layer is attached to a normalized layer in the output end.

\subsubsection{Encoder Layer}

This layer is a gated recurrent unit (GRU) layer used to encode the dependencies between the successive coordinates of the historical trajectories. In MATLAB~\textregistered, it is created using function \textit{gruLayer} with layer size $64$. This layer is also attached to a normalized layer in the output end.

\subsubsection{Latent Feature Layer}

This layer is a fully connected layer used to automatically extract the features from the encoded sequential data. It is created using function \textit{fullyConnectedLayer} with layer size $64$. Its output passes through a layer of Linear rectification functions (ReLU).

\subsubsection{Decoder Layer}

Similar to the encoder layer, this layer is also a GRU layer. It is used to decode the sequential features to sequential data that are used to generate the prediction. Its size is set as $128$.

\subsubsection{Output Layer}

This layer is used to map the decoded sequential data to the predicted trajectories. It is constructed by a fully connected layer sized 2 and a regression layer.

\section{Simulation Studies}\label{simulations}

In this section, we use a two-lane straight highway scenario to evaluate the MPC incorporating the predicted TV trajectory using deep learning. The scenario considers three parallel horizontal lanes, where the EV and the TV start at the middle and the bottom lanes, respectively. The EV is required to drive in the middle lane at a constant speed, and the TV needs to change to the middle lane, thus leading to possible collisions. We first evaluate the prediction precision of the trained RNN model. Then, we validate the efficacy of the MPC with predicted TV trajectory.

\subsection{Evaluation of the Prediction Model}
In this subsection, we evaluate the prediction accuracy of the RNN model. The loss function is based on mean squared errors (MSE) between the outputs of the RNN and the ground truth labels of the samples. Specifically, subtraction is performed between them in an element-wise manner. Then, the squared element-wise errors are summed up before being divided by the total number of elements. The training data set that contains $3973$ samples is used to train the model. The optimization of the MSE loss is solved using the Adam optimizer~\cite{kingma2014adam}. The training is performed on a Thinkpad laptop with Intel(R) Core(TM) i7-10750H CPU at $2.60\,$GHz. The entire training process takes $30$ epochs and $900$ iterations with a learning rate of $0.01$. We do not need a large number of epochs since predicting trajectories is not a heavy job. 

\begin{figure}[htbp]
    \centering
    \includegraphics[width=0.48\textwidth]{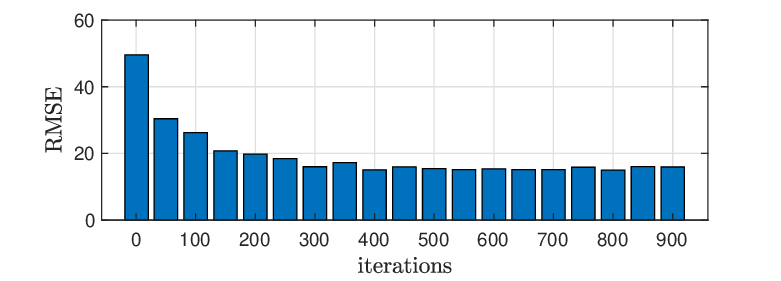}
    \caption{The training performance of the RNN model: the convergence of RMSE as the iteration increases.}
    \label{fig:training}
\end{figure}

The rooted MSEs (RMSE) as the iteration number increases are illustrated in Fig.~\ref{fig:training}. It is noticed that the RMSE decreases as the training proceeds with an ultimate score of $15.92$. The RMSE score becomes stable at around iteration $300$, which indicates the quick learning speed of the RNN model. This also reflects that trajectory prediction is an easy job for an RNN.

\begin{figure}[htbp]
    \centering
    \includegraphics[width=0.48\textwidth]{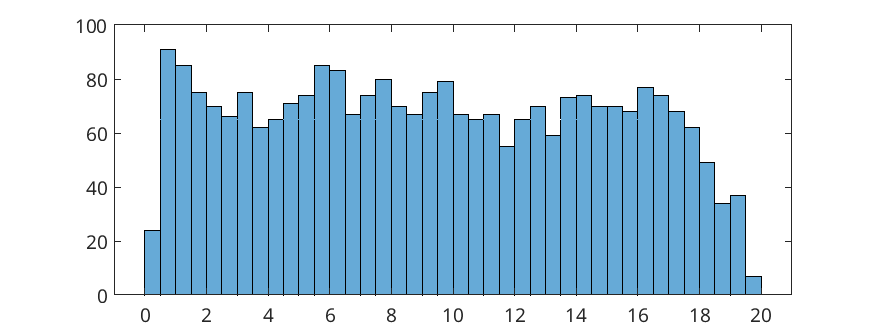}
    \caption{The training performance of the RNN model: the convergence of RMSE as the iteration increases.}
    \label{fig:hist}
\end{figure}

Then, we test the prediction accuracy of the trained RNN model using the test data set that contains $2649$ samples. We calculate the RMSE score for each predicted sample. The RMSE scores of all test samples are shown in the histogram chart in Fig.~\ref{fig:hist}. It can be seen that the RMSE scores of the prediction vary from $0$ to $20$. This range is very close to the ultimate training score of the model, $15.92$. The overall RMSE of the test is $10.91$. This indicates the accuracy of the trained RNN model for trajectory prediction.

\subsection{Incorporating the Prediction Model to MPC}

We use the predicted TV trajectory provided by the trained RNN model to generate the TV future trajectory ($x_k^{\mathrm{TV}}, y_k^{\mathrm{TV}}$), $k=0,1,\cdots,N$, and encode it to the safe set \eqref{safecon_exp}. Note that the starting point of the predicted TV trajectory may not be aligned with the current position of the TV due to the prediction error of the RNN model. This misalignment, however, can be eliminated by adding an offset to the predicted trajectory such that its starting point matches the current position of the TV.

The initial state of the EV is $\pmb{\xi}_0^\text{EV} = {[28,7.875,0,20]}^\intercal$. The EV is intended to reach a reference speed $20\ \text{m/s}$.
Then, we design an MPC for the EV as \eqref{eq:mpc} incorporating the interaction safety constraint \eqref{safecon_exp}. 
The parameters of the MPC are set as $N=10$, $T=0.2\,$s, $y \in [l^{\text{veh}},3w^{\text{lane}}-l^{\text{veh}}]$, $\psi \in [-1.2,  1.2]\,$\text{rad}, $v \in [0, 70]\,$m/s, $a \in [-9, 6]\,$m/{s}$^2$ and $\delta \in [-0.52,  0.52]\,$rad. The parameters of the elliptical safety set $\Xi_k^{\mathrm{safe}}$ are determined as $a=7\,$m and  $b=2.2\,$m.
The weighting matrices of the cost function are $Q=\text{diag}(0, 0.1, 0.001, 1)$, $R=\text{diag}(3,0.5)$ and $S=\text{diag}(0, 0.1, 0.001, 1)$.
The initial state of the TV is $\pmb{\xi}_0^\text{TV} = {[36,2.625,0,18]}^\intercal$. The trajectory of the TV is generated using a similar MPC to the EV but with a reference speed $20\ \text{m/s}$, the center lane as the target lane, and without incorporating the safety constraint \eqref{safecon}.  

\begin{figure}[htbp]
    \centering
    \includegraphics[width=0.96\linewidth]{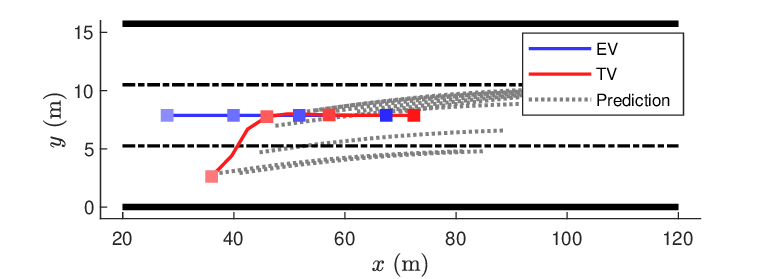}
    \caption{A two-vehicle lane-changing scenario.}
    \label{fig:DLresults}
\end{figure}

The generated trajectories of the EV and the TV within $11$ simulation steps are shown in Fig. \ref{fig:DLresults}, in blue and red, respectively. Their positions in the simulation steps 1, 4, 7, 11 are displayed as colored squares, shadow to dark in the order of time. The predicted TV trajectories at all simulation steps are also shown in the figure as dotted gray lines. From Fig. \ref{fig:DLresults}, we can see that the EV decelerates to avoid potential collisions with the TV as the TV starts to change its lane and then block the way of the EV in the center lane. When the TV finishes the lane-changing and accelerates for its own control target, the EV also starts to accelerate to reach its desired speed since it realizes that the risk of potential collisions is mitigated. This indicates that the EV is able to recognize the TV's lane-changing intention and be aware of the risk of collisions before the TV reaches the center lane. Therefore, the capability of the proposed MPC to mitigate the upcoming risks is addressed. Besides, from the figure, we can also see that the predicted TV trajectories become more and more consistent with the ground truth TV trajectory. This is because more data in the historical trajectory leads to higher prediction precision.

\section{Conclusion}\label{conclusion}
In this paper, we use deep learning to predict the trajectory of the TV and incorporate it into the safety constraint of the MPC of the EV. As a result, the EV is able to incorporate the risk of collisions with the TV into its motion planning such that it can make reasonable and risk-aware decisions in autonomous driving. Technical points such as data calibration and offset compensation are used to ensure that the MPC incorporates the more realistic predicted trajectory. An interesting case that is not investigated in this paper is that both vehicles can interact with each other, which renders a two-player game. Our method in this paper will be extended to this case in future work.

\addtolength{\textheight}{-12cm}   








\bibliographystyle{IEEEtran}
\bibliography{Reference}

\end{document}